\documentclass[letterpaper, 10 pt, conference]{ieeeconf}  
\usepackage{authblk}

\IEEEoverridecommandlockouts                              

\usepackage{amsmath,graphicx,amsfonts}
\usepackage{multirow}
\usepackage{caption, copyrightbox}

\usepackage{mwe} 

\title{\LARGE \bf A unified 3D framework for Organs at Risk Localization and Segmentation for Radiation Therapy Planning.}
\author[1,2,3,4]{Fernando Navarro$^{\dag}$ \thanks{\noindent ${\dag}$ equal contribution}}
\author[1]{Guido Sasahara$^{\dag}$}
\author[1,2]{Suprosanna Shit}
\author[1,2]{Ivan Ezhov}
\author[3]{Jan C. Peeken}
\author[3]{\\Stephanie E. Combs}
\author[4]{Bjoern H. Menze}
\affil[1]{Department of Informatics, Technical University of Munich, Germany}
\affil[2]{Center for Translational Cancer Research (TranslaTUM), Klinikum
rechts der Isar, Germany}
\affil[3]{Department of Radio Oncology and Radiation Therapy, Klinikum rechts der Isar, Munich, Germany}
\affil[4]{Department of Quantitative Biomedicine, University of Zurich, Zurich, Switzerland}

\begin{document}
%
\maketitle
\begin{abstract}
Automatic localization and segmentation of organs-at-risk (OAR) in CT are essential pre-processing steps in medical image analysis tasks, such as radiation therapy planning. For instance, the segmentation of OAR surrounding tumors enables the maximization of radiation to the tumor area without compromising the healthy tissues. However, the current medical workflow requires manual delineation of OAR, which is prone to errors and is annotator-dependent. In this work, we aim to introduce a unified 3D pipeline for OAR localization-segmentation rather than novel localization or segmentation architectures. To the best of our knowledge, our proposed framework fully enables the exploitation of 3D context information inherent in medical imaging. In the first step, a 3D multi-variate regression network predicts organs' centroids and bounding boxes. Secondly, 3D organ-specific segmentation networks are leveraged to generate a multi-organ segmentation map. Our method achieved an overall Dice score of $0.9260\pm 0.18 \%$ on the VISCERAL dataset containing CT scans with varying fields of view and multiple organs.
\end{abstract}
%
%



\section{Introduction}
Localization and segmentation of OAR are two necessary pre-processing steps for many automatic medical image analysis tasks, such as radiation therapy planning \cite{lipkova2019personalized,ezhov2020real} and lesion detection \cite{bilic2019liver}. Expert radiologists perform these steps manually, making the process time-consuming, costly, and dependent on the level of expertise. These problems call for computer-aided systems for automatic analysis. 

\begin{figure}[ht!]
  \centering
    \includegraphics[width=1\linewidth]{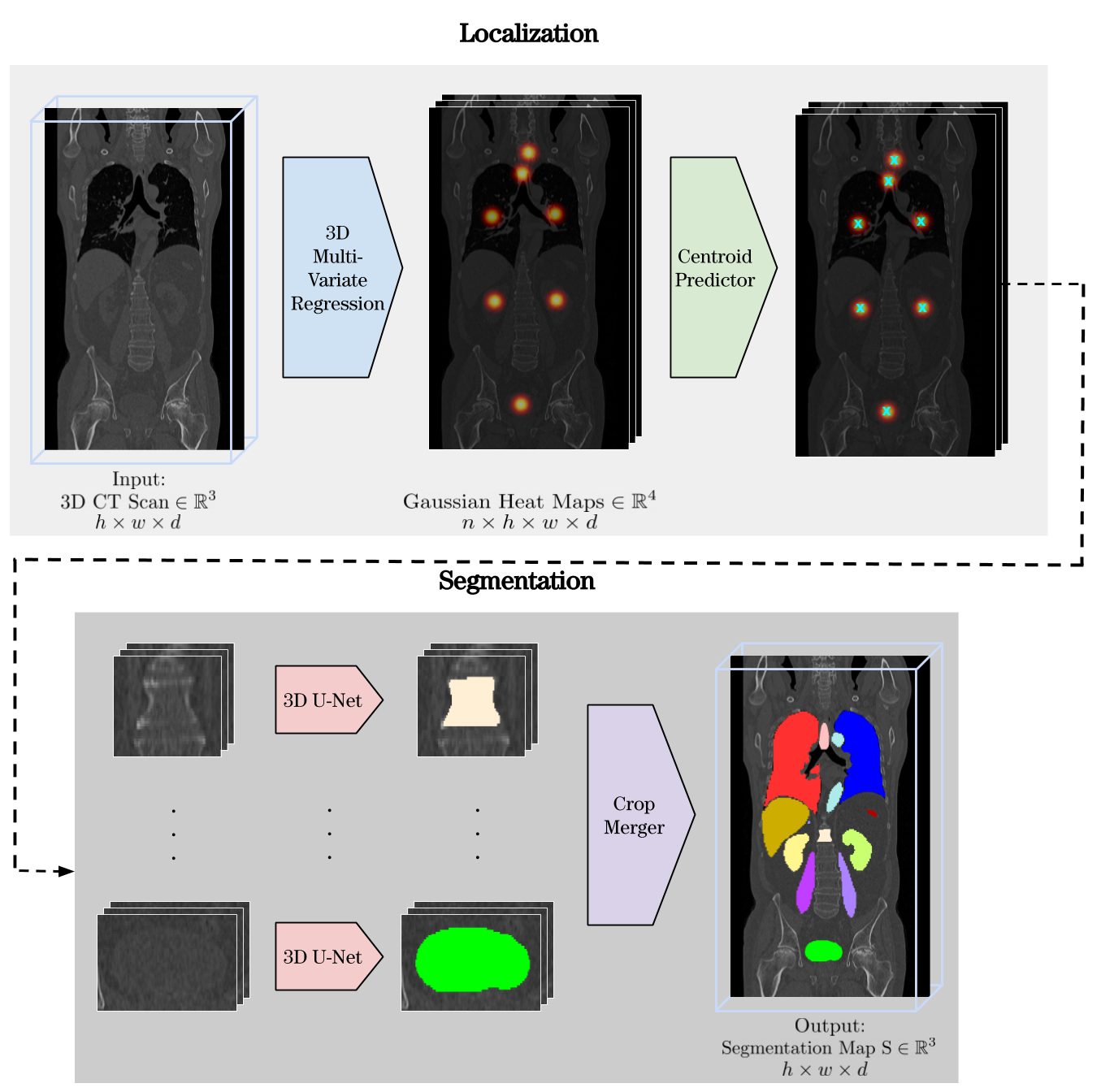}
    \caption{\small{Overview of the proposed approach. The input is a 3D CT scan. In the localization step, our proposed multi-variate regression network together with population statistics predicts organs' bounding boxes. In the segmentation step, segmentation is performed on organ-wise ROIs. The output of the network is the merged segmentation map after aggregation.}}
  \label{fig:overview}
\end{figure}
In recent years, deep convolutional neural networks have been widely adopted in medical imaging applications for both localization \cite{DeVos2017,ren2015faster,xu2019efficient,navarro2020deep,waldmannstetter2020reinforced} and segmentation \cite{ronneberger2015u,cciccek20163d,Navarro2019,qasim2020red}. These methods focus either on localization or segmentation. While the detection task prioritizes coarse-level organ representation to estimate relative position, the segmentation task requires fine feature representation to determine organ boundaries. We argue that in a holistic pipeline, prior knowledge about organ location obtained from the localization task is useful for the segmentation model to specialize in organ-specific feature learning solely. In this vein, few approaches aim at jointly solving the two tasks \cite{he2017mask,zhao2019knowledge, dabiri2020deep,hussain2021cascaded}.
Nevertheless, the model complexity of Mask-RCNN \cite{he2017mask} makes it unsuitable for 3D volumetric analysis. Likewise, methods similar to the proposed by Zhao et al. \cite{zhao2019knowledge} require a registration step resulting in a time-consuming approach. On the other hand, \cite{hussain2021cascaded, dabiri2020deep}  use 2D networks for localization using axial, coronal, and sagittal views without exploiting the 3D information. Other approaches like in \cite{vesal2020fully} segments the organ of interest twice: one in a coarse resolution using the gradcam for localization and later used a crop around the attention map to segment the organ. In \cite{liu2019automatic} patches generated from superpixels are classified as candidate regions of the organ of interest in the localization step followed by 2.5D segmentation networks. In contrast, our approach uses one network for centroids prediction and one for organs-specific segmentation.

The aforementioned two-step approaches are either too computational expensive for medical imaging or lack ability to fully exploit 3D information. In this work, we focus on bringing together 3D localization and segmentation pipelines to fully exploit 3D context information, rather than proposing novel localization or segmentation architectures. With this intention, we propose a holistic two-step approach to efficiently exploit the organ-size invariant relative positional information in a 3D context. Firstly, inspired by \cite{payer2020coarse}, we formulate the localization problem as a 3D multi-variate regression problem to predict the organs' centroids and hence the bounding boxes. Secondly, we leverage 3D segmentation networks to obtain the final segmentation map around the localized region obtained from the first step. Lastly, we aggregate each organ segmentation, resulting in a multi-organ segmentation map.

\section{METHODOLOGY}
\label{chapter:methodology}
Our proposed holistic approach consists of two stages: the localization and the segmentation stage. We obtain the final segmentation map by aggregating the individual segmentation predictions. Fig \ref{fig:overview}. shows an overview of the proposed approach. In the rest of the section, we describe the details of each component.


\subsection{Localization}

\subsubsection{\textbf{Multi-variate Gaussian Regression}}
We formulate the organ localization task as a multi-variate Gaussian regression problem for organs' centroids. These predictions in combination with average organs' bounding boxes enables the exploitation of 3D context information. Formally, we define the input 3D medical image as $\mathbf{X} \in \mathbb{R}^{(h \times w \times d)}$. For each organ, we define the ground truth as a Gaussian heat map of the form 
$\mathbf{y_i}= -e^{{\parallel x- \mu_{i} \parallel}^{2} /\ 2 \sigma^2}$, where $\mu_{i}$ is the centroid of the $i^{th}$ organ and $\sigma^2$ controls the variance of the Gaussian. The ground truth annotation results in $\mathbf{Y} \in \mathbb{R}^{(h \times w \times d \times n)}$, where $n$ is the number of organs to be localized.

\begin{figure}[h!]
  \centering
    \includegraphics[width=1.0\linewidth]{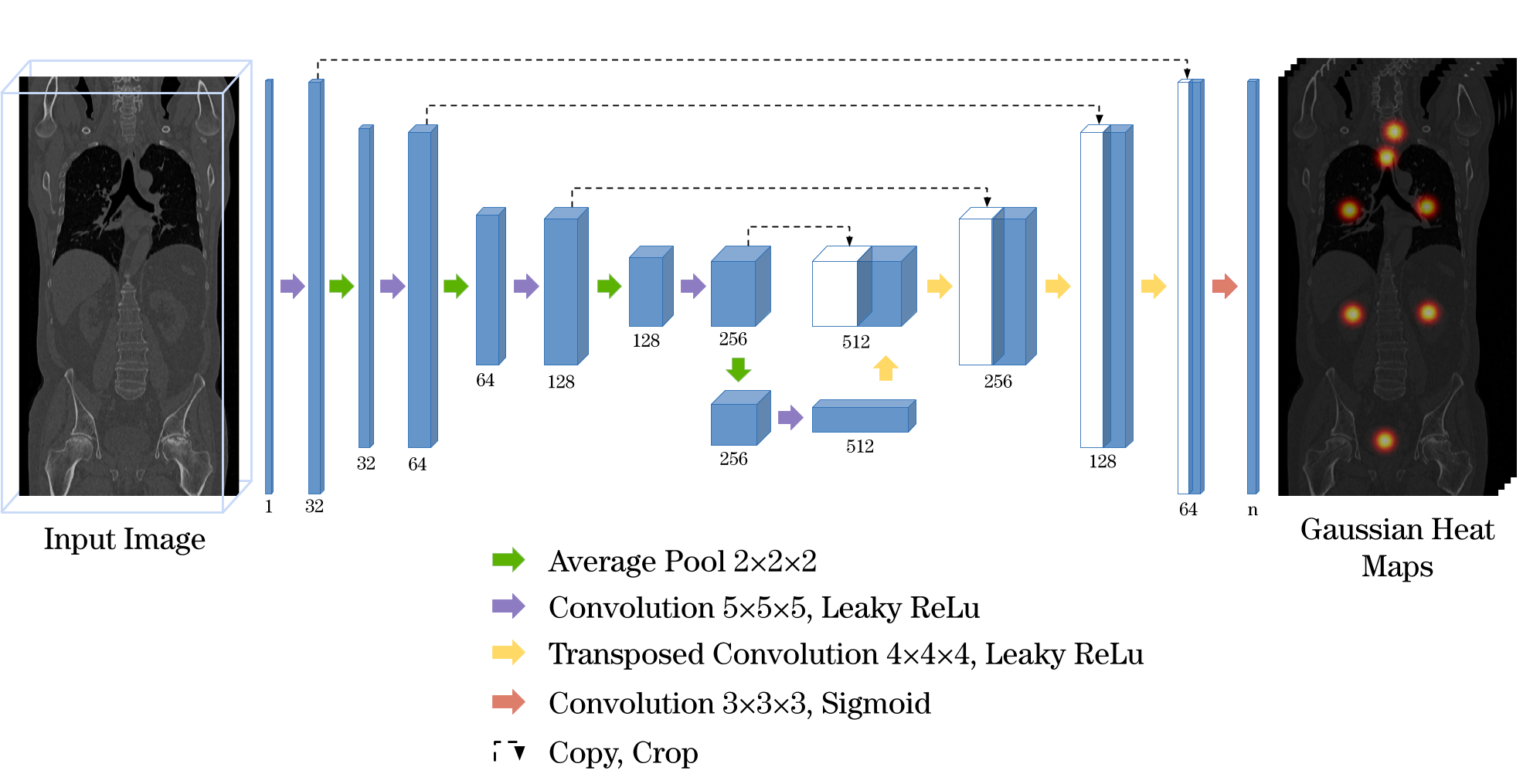}
    \caption{\small{Proposed 3D Multi-variate Gaussian regression network. Each output channel corresponds to one organ heat map.}}
  \label{fig:architecture}
\end{figure}

The proposed network architecture is inspired by the seminal works in Sekuboyina et al. \cite{Sekuboyina2018,sekuboyina2020labeling,payer2020coarse} for vertebrae localization and segmentation. We created a 3D version of the authors' approach \cite{Sekuboyina2018,sekuboyina2020labeling} to work directly in 3D volumetric data, instead of 2D projections, depicted in \textbf{ Fig.} \ref{fig:architecture}. In contrast to \cite{payer2020coarse}, we tailored the 3D localization proposed for vertebrae and adapted it to organs. This architecture allows the network to exploit the 3D context information and efficiently use computational resources for different CT organs. The loss function used for optimization is the $\mathcal{L}_{2}$ loss:
\begin{equation}
\mathcal{L}_{2} = \frac{1}{h\times w\times d \times n} \sum_{} {\parallel \mathbf{Y} - \tilde{\mathbf{Y}} \parallel}^2
\end{equation}
where $\mathbf{\tilde{Y}}$ stands for the predicted Gaussian heat map. 



\subsubsection{\textbf{Bounding Box Generation}}
To obtain bounding boxes from the predictions of the multi-variate Gaussian regression model $\mathbf{\tilde{Y}} \in \mathbb{R}^{(h \times w \times d \times n)}$, the following steps are executed:
\begin{itemize}
\renewcommand{\labelitemi}{$\bullet$}
  \item Voxels with Gaussian heat map's values above a threshold $\tau$ form a region potentially containing the organ centroid. This method is preferred over taking the highest heat-map probability as it is sensitive to outliers and false positives.
  \item We define the mid-point of this region as the organ centroid prediction.
  \item For each organ, an average bounding box size is computed from the training set.
  \item Finally, this bounding box is centered around the aforementioned centroid prediction plus $v$ voxels in each direction resulting in the final box. Adding this margin ensures the organ is contained inside the box.
\end{itemize}

\subsection{Organ-wise 3D Segmentation}

From the predicted organ's bounding boxes, we crop organ-wise ROIs from the CT scans. Then, we use these ROIs to train organ-specific 3D segmentation networks. This model segregation enables us to customize the networks to learn organ-specific feature representations. Furthermore, the localization stage endows the networks to process high-resolution 3D information efficiently. In this work, we adopted 3D U-Nets \cite{cciccek20163d} to generate the organ segmentation maps. 


We use a combination of binary cross-entropy and Dice as loss function:
\begin{equation}
\mathcal{L} = \underbrace{- \sum_{x}^{} {g_{}{(x)} \log{p_{}{(x)}}}}_\text{Cross-Entropy Loss} - \underbrace{{\frac{2 \sum_{x}^{} {p_{}{(x)}} g_{}{(x)}} {\sum_{x}^{} {p_{}^{2}{(x)}} + \sum_{x}^{} {g_{}^{2}{(x)}} } }}_\text{Dice Loss}
\end{equation}
where $p(x)$ is the probability of a pixel in location $x$ to belong to the foreground class and $g(x)$ is the ground truth for the corresponding pixel.

\begin{figure*}[h]
  \centering
    \includegraphics[width=0.7\linewidth]{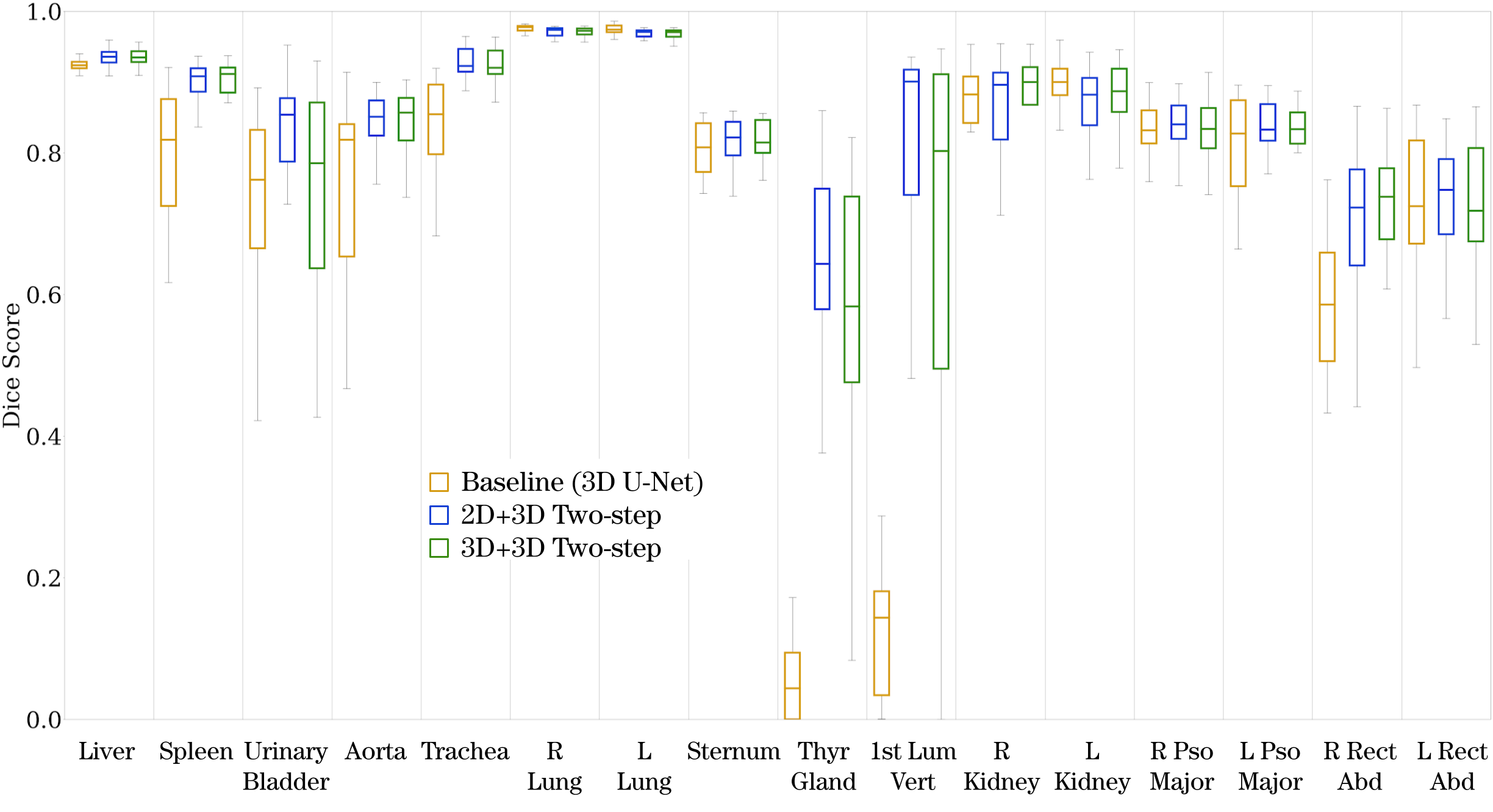}
    \caption{\small{Organ-wise box plots comparison across all tested methods. A constant improvement of  Dice score is obtained with the proposed approach compared to the baseline.}}
  \label{fig:dice_scores}
\end{figure*}
We aggregate the segmented crops from every organ to generate the final segmentation map, bringing the crops back to the original image space. During this aggregation process, there may exist overlap between organs that are close to each other. Those overlapping voxels, for which segmentation networks disagree, are set to the predictions with the highest probability value.



\section{Experiments and Discussion}\label{chapter:experiments}

To validate the performance of the proposed approach, we perform the segmentation of 16 anatomical structures considered as OAR: lungs, kidneys, liver, spleen, pancreas, aorta, urinary bladder, sternum, thyroid, vertebra, right-, and left-psoas and rectus abdominal muscles.


\textbf{Data set:} The data set used in the experiments consists of CT scans from the gold corpus, and silver corpus in the VISCERAL dataset \cite{Jimenez2016}. 74 CT scans from the silver corpus data are used for training and hyper-parameter tuning. The annotations in this set are automatically labeled by fusing the results of multiple algorithms. 23 CT scans from the gold corpus, which contains manually annotated labels, are used for testing. 




All volumes are resampled to $3mm^3$ and $1mm^3$ for the localization and segmentation stages, respectively. We normalized each volume to zero mean and unit variance.


To evaluate the overall performance of the proposed two-stage approach for multi-organ segmentation, we report the average dice score between the ground truth and the predicted segmentation map. In addition, organ-wise dice scores are reported using box plots.

\textbf{Optimization:} We train the networks using Adam optimizer with a decaying learning rate initialized at $1e^{-3}$ and $1e^{-5}$ for the multi-variate regression model and segmentation model, respectively. We use a Mini-batch size of 1 in all experiments. The training was continued till validation loss converged. All the experiments are conducted on an NVIDIA Titan Xp GPU with 12GB vRAM.


\textbf{Gaussian Variance:} For the generation of ground-truth Gaussian heat maps, several approaches were tested. Experimental results showed that an isotropic variance of \(150mm^2\) gives the best performance. However, larger variances resulted in less precise predictions, while smaller variances tended to cause localization failures, where no strong centroid prediction was generated. Organ-size-specific heat-maps had the tendency to suffer from the same pitfalls, being either too large or too small.


\textbf{Threshold for Centroid Prediction:} We found an empirical value of $\tau = 0.1$ to obtain the organ centroid prediction using the validation set. Furthermore, we found that heat maps' values  $<0.1$ translate to the absence of organ or detection failure.

\textbf{Quantitative Evaluation:} For the quantitative evaluation, we report our results and compare them with two approaches. Here is the definition of the experiments:
\begin{itemize}
\renewcommand{\labelitemi}{$\bullet$}
  \item \textbf{Baseline (3D U-Net)}: segmentation directly in 3D, this is a 3D version of the original U-Net with weighted cross-entropy for class imbalance.
  \item \textbf{2D+3D Two-step}: 2D localization proposed in \cite{DeVos2017} followed by 3D U-Net segmentation.
  \item \textbf{3D+3D Two-step}: Proposed 3D localization followed by 3D U-Net segmentation.
\end{itemize}


In \textbf{Table} \ref{tab:total_dice} we report the overall Dice score for every method. We can observe that our proposed approach results in an improvement of $1.4\%$ compared to the baseline. Similarly, the 2D+3D Two-step approach results in a similar performance to our proposed approach. Nevertheless, notice that the 2D localization proposed by DeVos et al. \cite{DeVos2017} requires the training of three independent 2D networks in axial, coronal, and sagittal views to generate the final bounding box, while our approach only requires one network.

\begin{table}[h]
\begin{center}
\begin{tabular}{ll}
\textbf{Method}                              & \textbf{Dice Score} \\
\noalign{\hrule height 0.8pt}
Baseline (3D U-Net)                           & 0.9120 \(\pm\) 0.026                        \\ \hline
2D+3D Two-step            & \textbf{0.9274 \(\pm\) 0.016 }                       \\ \hline
3D+3D Two-step & \textbf{0.9260 \(\pm\)  0.018}                       \\ \hline
\end{tabular}
\caption{ \small{Quantitative results: the table describes the mean and standard deviation of the global Dice scores for all evaluated models.}}
    \label{tab:total_dice}
\end{center}
\end{table}
In \textbf{Fig.} \ref{fig:dice_scores} we can observe the organ-wise performance across different evaluated methods using box plots. We can notice that Dice scores for big structures, for instance, lungs and liver, do not improve significantly compared to the baseline. In contrast, the proposed two-step approach provides significant improvement for small structures such as the spleen, urinary bladder, aorta, trachea, thyroid gland, first lumbar vertebra, and muscles. We attribute this to the fact that training a network to segment all structures at once creates class imbalance and small structures are underrepresented. The proposed two-step approach resolves this problem, delivering better performance for these structures.

\begin{figure}[h]
  \centering
    \includegraphics[width=1.0\linewidth]{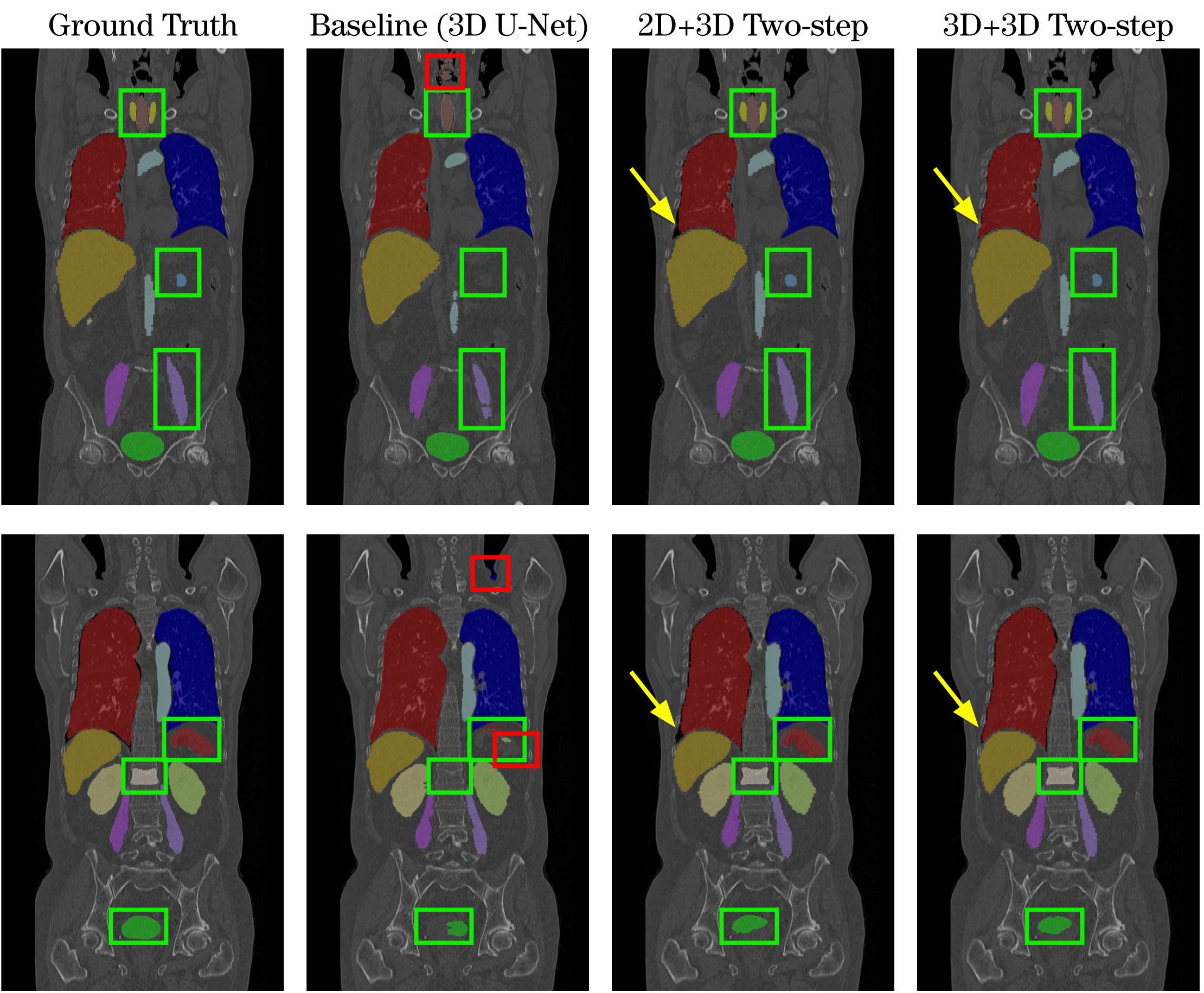}
    \caption{\small{Qualitative results: ROIs in green indicate regions where the proposed approach improves the segmentation. ROIs in red indicate regions where the baseline produces false positives. Yellow arrows point to regions where the proposed approach presents smooth and continuous segmentations.}}
  \label{fig:segimg}
\end{figure}

\textbf{Qualitative Evaluation:} In \textbf{Fig} \ref{fig:segimg}  we show a qualitative comparison between the baseline, the two-step approach with 2D localization, and the proposed approach. We highlight ROIs with green boxes to show regions where our approach improves the segmentation and red boxes to indicate regions where baseline produces false positives. We can observe the correlation of Dice score improvement in the quantitative results with the visual assessment, particularly for the small organs. For instance, the highlighted red boxes show examples of false positives for the trachea and lungs. Moreover, observing the last two columns of \textbf{Fig} \ref{fig:segimg}, we can visually recognize that the proposed approach results in more smooth and continuous segmentation compared to both the baselines and 2D+3D approach. This is clearly observed on the boundaries of the lungs' segmentations indicated with yellow arrows in \textbf{Fig.} \ref{fig:segimg}.

\section{Conclusion}

In this work, we introduce a unified 3D pipeline for OAR localization-segmentation rather than novel localization or segmentation architectures. The proposed localization framework and the  unified two-step approach for multi-organ segmentation work directly in 3D volumetric data to exploit 3D context information. We have validated our method in a public benchmark data set, which shows consistent improvement in Dice score compared to the baseline, especially for the small structures. Future work includes directly predicting the extension of the bounding boxes along the centroid prediction to discard the need for population statistics.




\bibliographystyle{IEEEbib}
\bibliography{refs}

\end{document}